\documentclass{IEEEoj}
\usepackage{cite}
\usepackage{amsmath,amssymb,amsfonts}
\usepackage{algorithmic}
\usepackage[pdftex]{graphicx}
\usepackage{textcomp}
\usepackage[ruled,vlined]{algorithm2e}
\usepackage[export]{adjustbox}
\usepackage{epstopdf}
\DeclareGraphicsExtensions{.eps,.ps,.pdf,.png,.jpg}

\usepackage{tikz}
\usepackage{pgfplots}
\pgfplotsset{compat=1.17}

\usepackage{hyperref}
\hypersetup{
	colorlinks=true,
	linkcolor=black,     
	urlcolor=blue,
	citecolor=black,
}
\SetAlFnt{\small}

\definecolor{commentcolor}{RGB}{60,124,125}   
\newcommand{\PyComment}[1]{\ttfamily\textbf{\textcolor{commentcolor}{\# #1}}}  
\newcommand{\PyCode}[1]{\ttfamily\textbf{\textcolor{black}{#1}}} 
\DeclareMathOperator*{\argmin}{argmin}

\def\BibTeX{{\rm B\kern-.05em{\sc i\kern-.025em b}\kern-.08em
    T\kern-.1667em\lower.7ex\hbox{E}\kern-.125emX}}
\AtBeginDocument{\definecolor{ojcolor}{cmyk}{0.93,0.59,0.15,0.02}}
\def\OJlogo{\vspace{-14pt}\includegraphics[height=28pt]{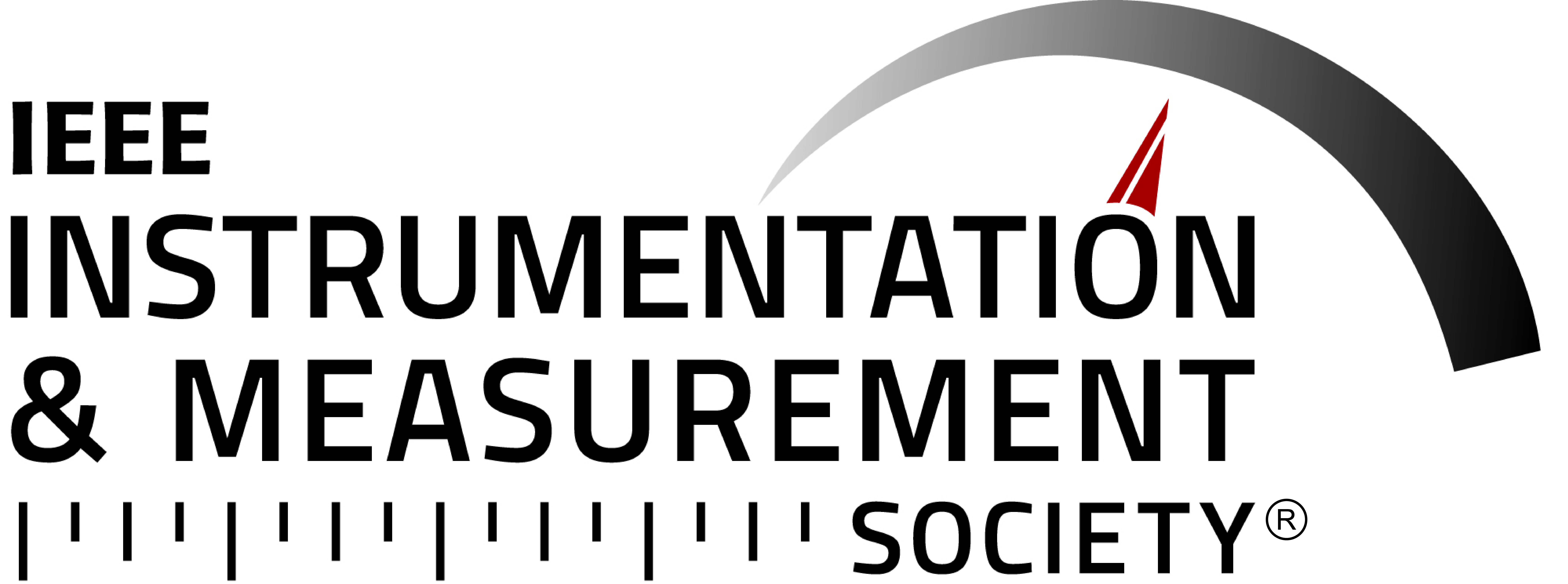}}

\begin{document}
\receiveddate{XX Month, XXXX}
\reviseddate{XX Month, XXXX}
\accepteddate{XX Month, XXXX}
\publisheddate{XX Month, XXXX}
\currentdate{XX Month, XXXX}
\doiinfo{OJIM.2022.1234567}

\DeclareRobustCommand*{\IEEEauthorrefmark}[1]{%
  \raisebox{0pt}[0pt][0pt]{\textsuperscript{\footnotesize\ensuremath{#1}}}}

\title{Deep Learning-Based Anomaly Detection in Synthetic Aperture Radar Imaging}

\author{Max Muzeau\authorrefmark{1,2}, Chengfang Ren\authorrefmark{2}, Member, IEEE, S\'ebastien Angelliaume\authorrefmark{2}, Mihai Datcu\authorrefmark{3}, Fellow, IEEE, and Jean-Philippe Ovarlez\authorrefmark{1,2}, Member, IEEE}
\affil{SONDRA, CentraleSup\'elec, Universit\'e Paris-Saclay, France}
\affil{DEMR, ONERA, Universit\'e Paris-Saclay, France} \affil{University Politehnica of Bucharest (UPB), Romania and German Aerospace Center (DLR), Germany}
\corresp{Corresponding Author: M. Muzeau (e-mail: max.muzeau@ onera.fr).}
\markboth{Preparation of Papers for IEEE OPEN JOURNALS}{Author \textit{et al.}}

\begin{abstract}
In this paper, we proposed to investigate unsupervised anomaly detection in Synthetic Aperture Radar (SAR) images. Our approach considers anomalies as abnormal patterns that deviate from their surroundings but without any prior knowledge of their characteristics. In the literature, most model-based algorithms face three main issues. First, the speckle noise corrupts the image and potentially leads to numerous false detections. 
Second, statistical approaches may exhibit deficiencies in modeling spatial correlation in SAR images. Finally, neural networks based on supervised learning approaches are not recommended due to the lack of annotated SAR data, notably for the class of abnormal patterns. Our proposed method aims to address these issues through a self-supervised algorithm. The speckle is first removed through the deep learning SAR2SAR algorithm. Then, an adversarial autoencoder is trained to reconstruct an anomaly-free SAR image. Finally, a change detection processing step is applied between the input and the output to detect anomalies. Experiments are performed to show the advantages of our method compared to the conventional Reed-Xiaoli algorithm, highlighting the importance of an efficient despeckling pre-processing step.
\end{abstract}


\maketitle

\section{INTRODUCTION}

Anomaly detection is one of the most critical issues in multidimensional imaging, mainly in hyperspectral \cite{Stein02,Matteoli10} and medical imaging . Even if we have no prior information about a target or background signature, anomalies generally differ from surrounding pixels due to their dissimilar signatures. Anomaly detection for radar and SAR imaging aims to discover abnormal patterns hidden in multidimensional radar signals and images. Such anomalies could be man-made changes in a specific location or natural processes affecting a particular area. These anomalies can characterize, for example, several potential applications: Oil slick detection \cite{Alpers17}, turbulent ship wake \cite{Graziano17}, levee anomaly \cite{Fisher17} or archaeology \cite{Scollar90}. They can also be related to any change detection in time-series SAR images. This research field is essential in data mining for quickly isolating irregular or suspicious segments in large amounts of the database. Many anomaly detection schemes have been proposed in the literature \cite{Reed90, Frontera14, Veganzones14, Terreaux15, Frontera16, Bitar19, Haitman19, Akcay19, Schlegl19, Sinha20, Mabu21}. Among them, the unsupervised methods are the most interesting since they are widely applicable and do not require labeling the data.

Deep learning techniques for anomaly detection are often based on an encoding-decoding network architecture to learn healthy data or images that do not contain anomalies, such as references \cite{Akcay19, Schlegl19}. 
Another approach is based on the one-class SVM method in the latent space \cite{Mabu21}. 
Anomaly detection in SAR using the difference between the input and the reconstructed image obtained through the autoencoder scheme is also proposed in \cite{Sinha20}. Still, it could suffer from the SAR speckle noise. 

Our approach consists in mitigating this annoying speckle before training an Adversarial Autoencoder (AAE) from these despeckled data. It does not need to separate abnormal SAR images from the training dataset.We then compare our method with the conventional RX anomaly detector \cite{Reed90} by analyzing their global performance, and we highlight the importance of the despeckling process.  

This paper is organized as follows. After some introduction in Section I, Section II introduces the context of SAR imaging. The proposed methodology being based on despeckling of SAR images, Section III describes the filtering process. Section IV presents the self-supervised neural architecture that aims to reconstruct the final SAR image without anomalies. Section V will define the last phase of the processing based on the change detection step and will discuss the quality of the anomaly detection procedure. The proposed methodology is applied to experimental polarimetric SAR images. Final Section VI gives some conclusions and perspectives.

 \begin{figure*}[t]
   \centering
   \includegraphics[width=\linewidth]{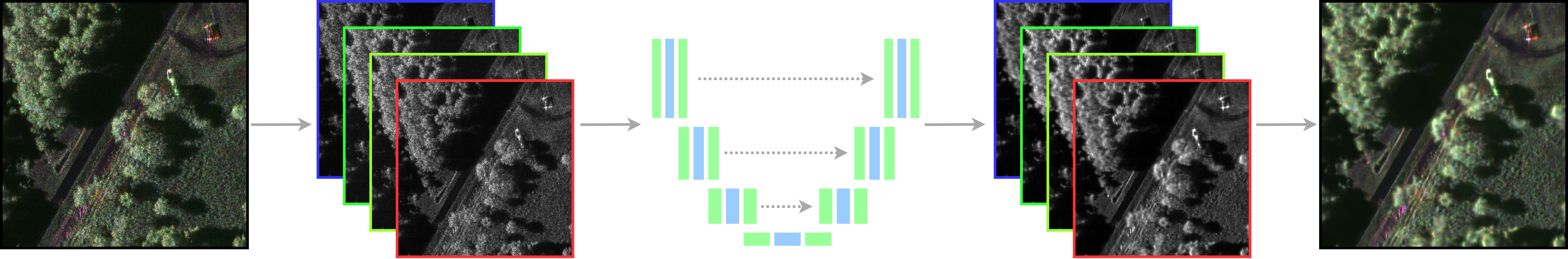}  
  \caption{Despeckling method with the architecture described in \cite{SAR2SAR}. First, we decompose a polarimetric SAR image in four different intensity images (discarding the phase). Each polarization is then denoised separately, and the polarimetric SAR image is finally reconstructed.}
  \label{fig:SAR2SAR}
\end{figure*}

\section{SYNTHETIC APERTURE RADAR}
\label{section:SAR}
Airborne and spaceborne SAR aims to provide images of Earth's surface at radar frequency bands. Contrary to optical imaging, they can work day and night using an active radar that transmits and receive pulses in the scanning area. They offer an opportunity to monitor changes and anomalies on the Earth's surface. Their principle is to combine multiple received signals coherently to simulate an antenna with a larger aperture. This procedure allows building a complex-valued image of the terrain with very high range and azimuth resolutions, independently of the distance between the radar and the imaged area. Electromagnetic waves can also be polarized at emission and reception. Horizontal and vertical polarizations are generally used, resulting in four coherent SAR channels ($HH,HV,VH,VV$). $HV$ corresponds to a horizontal polarization for the emission and vertical polarization for the reception. The information provided by polarimetry is crucial for most geoscience applications from SAR remote sensing data \cite{Pottier09}.

\subsection{Speckle}
The main issue in SAR imaging concerns the corruption of the backscattered signal within a resolution cell by a multiplicative noise called speckle. This is due to the coherent summation of multiple scatterers in one resolution cell, which may cause destructive or constructive inferences. This phenomenon disturbs the exploitation of radar data  for detection and geoscience applications. Much work in the literature has proposed reducing its effect (e.g. \cite {Deledalle09, Deledalle11, Deledalle15, Deledalle17, SAR2SAR, Dalsasso22a, Dalsasso22b}) while preserving the resolution as much as possible.

A model of the speckle has been defined by Goodman \cite{Goodman76}. In the case of a single look complex (SLC) image, each pixel power or intensity $I$ may be distributed according to an exponential distribution:
\begin{equation}
\label{eq:1}
p(I|R) = \frac{1}{R} \, \exp{\left(-\frac{I}{R}\right)}\, ,
\end{equation}
where $R$ denotes the mean reflectivity level. A useful general statistical model representation is the so-called multiplicative noise model:
\begin{equation}
\label{eq:2}
I = R \, S\, ,
\end{equation}
where $S$ characterizes the exponentially distributed speckle and $R$ is a scalar positive random variable characterizing the texture. 

To change the multiplicative noise model into an additive one, a log transformation is often used, which leads to a new distribution:
\begin{equation}
\label{eq:3}
p(y|x) = \exp{(y-x)} \, \exp(-\exp{(y-x)})\, ,
\end{equation}
where $y = \log(I)$ and $x = \log(R)$.
 One crucial detail is the spatial correlation between the pixels which is not always taken into account in the above model. This can be mainly explained by a possible apodization of the images or some oversampling lower than the spatial resolution during the image synthesis.
 We can directly work on full-resolution data thanks to recent developments in deep learning despeckling algorithms that know how to keep the spatial resolutions and preserve details such as lines and small bright targets (boats or vehicles, for example). 
 They, therefore, render detection tasks and false alarm regulation easier on speckle-free data. 

 \subsection{Dataset}
 \label{subsection:data}
In this paper, the analyzed dataset is composed of a time series of X-band SAR images (each of size $4800\times 30000\times 4$) acquired by SETHI, the airborne instrument developed by ONERA  \cite{SETHI19, Angelliaume17}. The resolutions of these images are about $20$ cm in both azimuth and range domains for the four polarization channels ($HH,HV,VH,VV$).

In the monostatic case, the channels $VH$ and $HV$ are often averaged because they contain the same information. The averaging decreases the speckle impact without degrading the resolution (so-called reciprocity principle \cite{Palotta20}). The resulting three channels $[HH,\frac{1}{2}(HV+VH),VV]$ are then thresholded separately using a value $\lambda(I)$ chosen as:
\begin{equation}
\label{eq:4}
\lambda(I) = \mu_I + 3\,\sigma_I\, ,
\end{equation}
where $\mu_I$ and $\sigma_I$ are respectively the estimated mean and the standard deviation of the image $I$. The threshold  $\lambda(I)$ is the same for all SAR images to have a consistent visualization, as shown in Fig.~\ref{fig:SAR2SAR}. 

\section{Despeckling}
\label{sec:Despeckling}

To reduce the speckle impact, the deep-learning despeckling algorithm SAR2SAR \cite{SAR2SAR} has been used. As discussed previously, this method can estimate the spatial correlation between pixels and thus works with full-resolution images.

The training phase is based on a method called {\it Noise2Noise} first developed in \cite{Noise2Noise18}. It shows that there is no need for a ground truth to train a denoising deep neural network, but only for multiple images of the same area with different noise realizations. A common strategy for estimating the true unknown image is to find, for the quadratic $L^2$ loss function, an image $\bf{z}$ that has the smallest average deviation from the measurements:
\begin{equation}
\label{eq:5}
\argmin_\mathbf{z} \mathbb{E}_\mathbf{y} \big[ \left\lVert\mathbf{z}-\mathbf{y}\right\rVert^2 \big] = \mathbb{E}_\mathbf{y}[ \mathbf{y} ]\, ,
\end{equation}
where $\mathbf{y}$ represents a set of observations. If the sample noise is additive, centered, and independent in all the observations, estimating $\mathbb{E}_\mathbf{y}[ \mathbf{y}]$ is the same as predicting noise-free observations. 

The same principle has been applied and refined for SAR images in the algorithm \cite{SAR2SAR}: let $\mathbf{y}_1$ and $\mathbf{y}_2$ be two independent realizations of identically distributed random variables and let a denoising network $f_{\mathbf{\theta}}(.)$ be parameterized by $\mathbf{\theta}$. A loss $L_{speck}$ based on the negative log-likelihood of log-intensity images can be used according to the distribution defined in Eq.~\eqref{eq:3} and leads to:
\begin{equation}
\label{eq:6}
L_{speck} =  \sum_k f_{\mathbf{\theta}}([\mathbf{y}_1]_k) - [\mathbf{y}_2]_k + \exp\left([\mathbf{y}_2]_k - f_{\mathbf{\theta}}([\mathbf{y}_1]_k\right) 
\end{equation}

The loss is a sum of all partial loss for each pixel represented with the index $k$. The use of Eq.~\eqref{eq:6}, instead of a mean square error, allows a faster convergence.

The other difference between SAR2SAR and Noise2Noise is the training phase which is decomposed into three steps:
\begin{itemize}
    \item Phase A: The training step is done with a speckle-free image $\mathbf{y}$. Two fake noisy image are created such that $\mathbf{y}_1 = \mathbf{y}+\mathbf{s}_1 $ and $\mathbf{y}_2 = \mathbf{y}+\mathbf{s}_2 $ where $\mathbf{s}_1$ and $\mathbf{s}_2$ are both distributed according  the distribution from Eq.~\eqref{eq:3}. The network is learning how to denoise data without any knowledge of  spatial correlation.
    \item Phase B: The training step is performed with experimental SAR images and a change compensation pre-processing: two same areas acquired at a different time $\mathbf{y}_{t_n}$ and $\mathbf{y}_{t_m}$ are used. To compensate for changes that could have occurred between the acquisition at two dates, an estimation of the reflectivity is done beforehand (on sub-sampled images to remove spatial correlation). This leads to $\mathbf{y}_1 = \mathbf{y}_{t_n}$ and $\mathbf{y}_2 = \mathbf{y}_{t_m}-f_{\mathbf{\theta}}(\mathbf{y}_{t_m}^{sub})+f_{\mathbf{\theta}}(\mathbf{y}_{t_n}^{sub})$ where the $sub$ exponent characterizes sub-sampled data. This finally allows $\mathbf{y}_2$ to have the speckle of $\mathbf{y}_{t_m}$ and the reflectivity of $\mathbf{y}_{t_n}$.
    \item Phase C: The same operation is done, but this time, the network has learned how to model spatial correlation, so the images are not sub-sampled to estimate a reflectivity beforehand. This gives $\mathbf{y}_1 = \mathbf{y}_{t_n}  $ and $\mathbf{y}_2 = \mathbf{y}_{t_m}-f_{\mathbf{\theta}}(\mathbf{y}_{t_m})+f_{\mathbf{\theta}}(\mathbf{y}_{t_n})$.
\end{itemize}

\section{Image reconstruction}
Once the pre-processing step has been performed, the goal is to highlight anomalies. This is obtained in an unsupervised manner, and the network has no information about what should be an anomaly and what shouldn't. This makes the task harder than, for example, with a supervised convolution network. But, if we can reach the state-of-the-art with an unsupervised algorithm, one of the most significant drawbacks of artificial intelligence, corresponding to the need for a high-quality labeled dataset will be overcome.

\subsection{Proposed architecture}
We use an AAE \cite{Makhzani16} with convolution layers to deal with this problem. To understand the architecture, it is first necessary to introduce what is a Generative Adversarial Network (GAN) \cite{Goodfellow20}. We will use the notations $\mathcal{E}(.)$,  $\mathcal{D}(.)$ and $\mathcal{D}_c(.)$ to define the encoder, the decoder, and the discriminator respectively. A description of the network is illustrated in Fig.~\ref{fig:AAE}.

\vspace{-0.5cm}
\subsubsection{Generative adversarial networks}
A GAN is a self-supervised deep learning algorithm that was originally used to generate synthesized images based on what it saw. A generator and a discriminator are trained jointly against each other. The generator's goal is to create the most realistic image possible based on a source vector of low dimension. Usually, this vector is distributed according to a multivariate Normal distribution. The generated image should fool the discriminator into thinking there is no difference between real and fake images. In opposition, the discriminator's goal is to be able to distinguish fake generated data from real ones. In our architecture, the role of the GAN is to generate a latent vector that is distributed according to a Normal distribution representative of a patch of polarimetric SAR speckle-free image.
\vspace{-0.5cm}
\subsubsection{Adversarial Autoencoder}
An AAE is composed of two things:
\begin{itemize}
  \item A generator, which is an encoder and a decoder, one after another,
  \item A discriminator that is placed between the encoder and the decoder. His goal is to ensure that the latent space follows a normal distribution.
\end{itemize}

Our input is the logarithm of a denoised polarimetric SAR patch $\mathbf{X}\in \mathbb{R}^{h\times w\times c}$ of size $h \times w$ and depth $c$, where $h$ and $w$ are respectively the numbers of pixels in azimuth and range, and where $c$ is the polarization channel. The logarithm operation is used to reduce the dynamic of the data (for example, there can be a difference of $10^3$ between the amplitude of a strong scatterer and the amplitude of a pixel located in a vegetation area). After passing through the encoder, we get a latent vector $\mathbf{z} = \mathcal{E}(\mathbf{X})$ with $\mathbf{z}\sim p(\mathbf{z})$ and where $p(.)$ is the {\it a priori} distribution of our encoder. Based on this vector, the decoder will then make an estimation of our input patch $\hat{\mathbf{X}} = \mathcal{D}(\mathbf{z})$. In our architecture, we can observe a GAN present  with the generator $\mathcal{E}(.)$ and the discriminator  $\mathcal{D}_c(.)$ whose purpose is to differentiate $\mathbf{z}_{real} \sim \mathcal{N}(\mathbf{0},\mathbf{I})$ from $\mathbf{z}_{fake}\sim p(\mathbf{z})$. Here, the latent space is distributed according to a reduced and centered Normal distribution, but the Uniform distribution on [0,1] could also have given similar results. One restriction is that it could not have been a long tail distribution because the information would not have been compacted in a small enough space.
We then train the AAE in two successive phases:
\begin{itemize}
\item[1] Reconstruction error: this characterizes a loss that ensures to have a low pixel-per-pixel error. We minimize $L_{rec}$, which is a $L^1$ norm. It has been preferred instead of a $L^2$ norm because we do not want to heavily penalize the network when the difference between $\mathbf{X}$ and $\hat{\mathbf{X}}$ is huge.
\begin{equation}
\label{eq:7}
L_{rec} = \frac{1}{h\,  w\,  c}\displaystyle \sum\limits_{i,j,k} \big\lVert \mathbf{X}_{i,j,k} - \hat{\mathbf{X}}_{i,j,k} \big\rVert_1 \, .
\end{equation}
\item[2] Regularization error: such a loss allows us to control the distribution $p(\mathbf{z})$. It also gives a better reconstruction according to \cite{Mabu21,Makhzani16}. To do so, the weights of the encoder $\mathcal{E}$ and the discriminator $\mathcal{D}_c$ are estimated according to: 
\begin{eqnarray}
\label{eq:8}
\!\!\!\!\!\!\! \underset{\mathcal{E}}{\min} \, \underset{\mathcal{D}_c}{\max} \, L_{lat} \!\! \! &\!\!  \widehat{=} \!\!& \!\! \mathbb{E}_{\mathbf{z}_{real}\sim \mathcal{N}(\mathbf{0},\mathbf{I})}\left[\log(\mathcal{D}_c(\mathbf{z}_{real})\right]  \nonumber \\ 
\!\!\!\!& + & \!\! \mathbb{E}_{\mathcal{E}(\mathbf{X})\sim p(z)}\left[\log\left(1-\mathcal{D}_c(\mathcal{E}(\mathbf{X})\right)\right]\, .
\end{eqnarray}
\end{itemize}
\vspace{-0.5cm}
\subsubsection{Link between anomaly detection and AAE}
It is not apparent to see the link between this architecture and anomaly detection. The goal of an AAE for this application will be to make an accurate estimation for Normal distributed data and a bad estimation for abnormal data. This will be based on the assumption that the network is not powerful enough to reconstruct every aspect of the image. Only the recurrent patterns will be remembered. By definition, a spatial anomaly occurs rarely compared to the rest of the data. If all these assumptions are valid, only rare patterns will be seen as anomalies, which is exactly what we want. 

\begin{figure}
   \centering
   \includegraphics[width=\linewidth]{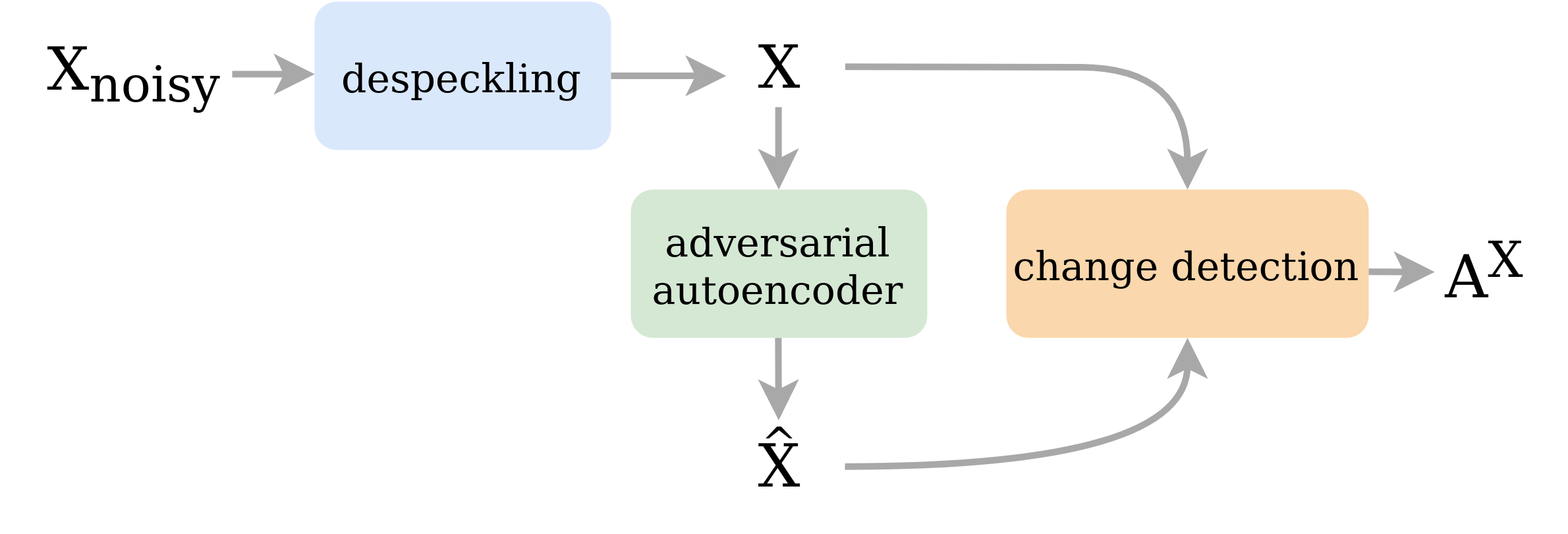} 
  \caption{Overview of our proposed architecture for SAR anomaly detection. $X_{noisy}$ corresponds to the amplitude of SAR image, $X$ is the speckle-free version, and $\hat{X}$ is its reconstruction using an adversarial autoencoder. Finally, the anomaly map $A^X$ is obtained through a change detection algorithm between $X$ and $\hat{X}$}
  \label{fig:archi}
\end{figure}

\section{Change detection method}
Once the network has delivered an estimation of the input image, the goal is to detect changes between $\mathbf{X}$ and $\hat{\mathbf{X}}$. This detection will be represented in an anomaly map $A^{\mathbf{X}}\in [0,1]^{h\times w}$. The closer the value is to one, the more the pixel is likely to be an anomaly. The global flow of the detection procedure is shown in Fig.~\ref{fig:archi}.

\subsection{Problem formulation}
There are multiple approaches to detect changes between two or more images \cite{Mian21}. As described in \cite{Mian21}, pixel-level comparisons are widely used in SAR imaging community. The problem is formulated within the set of the following two hypotheses:
\begin{equation}
\left\{
    \begin{array}{ll}
         H_0: \boldsymbol{\theta}_1 = \boldsymbol{\theta}_2 \text{ (no anomaly),}\\
         H_1: \boldsymbol{\theta}_1 \neq \boldsymbol{\theta}_2 \text{ (anomaly),}
    \end{array}
\right.
\end{equation}
where $\boldsymbol{\theta}_1$ and $\boldsymbol{\theta}_2$ are vectors of the estimated parameters of the distribution used to model pixel values.
Methods relying on the hypothesis test on the statistics of the image are difficult to exploit. The non-Gaussianity, the heterogeneity of the SAR images, or their complex-valued nature make this derivation very difficult. 

When pixels have been transformed into log intensity for better contrast and when the speckle effect has been reduced beforehand, the final distributions of the proposed anomaly detection tests are generally unknown. Hence, to be able to compare all the proposed strategies, a visualization process is used to threshold each map according to the following equation:  
\begin{equation}
\label{eq:PFA}
\Tilde{x} = \min(x,t)\, .
\end{equation}

For a given percentage $p$, the threshold $t$ is fixed such that $p\%$ of the pixels are above it. This ensures to have the same number of pixels of value $t$ for each result. The dynamic is compressed in a way that allows us to see at the same time the anomalies and the background. The Probability of False Alarm (PFA) is here, for convenience, characterized by this value $p$ in the sense that $p$ and PFA are equal if all the pixels in the map correspond to $H_0$ hypothesis. They will have the same significance in the sequel.  

One way to detect changes between two SAR images consists of testing the equality between the two estimated covariance matrices of the corresponding pixel under test for each pixel. This can be made statistically if the knowledge of the data statistic is known \cite{Novak05} or through matrix distances \cite{Forstner03}.  

\begin{figure}
\begin{minipage}[]{0.32\linewidth}
   \centering
   \includegraphics[width=\linewidth]{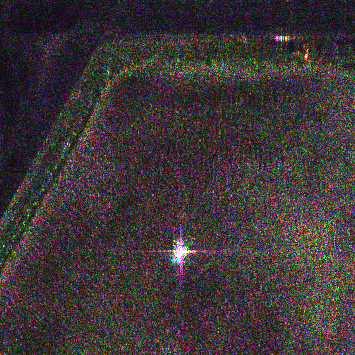}     
  \end{minipage}
    \hfill
  \begin{minipage}[]{0.32\linewidth}
   \centering
   \includegraphics[width=\linewidth]{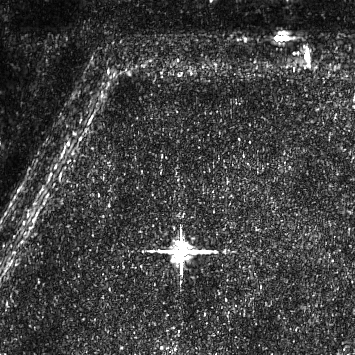}     
  \end{minipage}
  \hfill
  \begin{minipage}[]{0.32\linewidth}
   \centering
   \includegraphics[width=\linewidth]{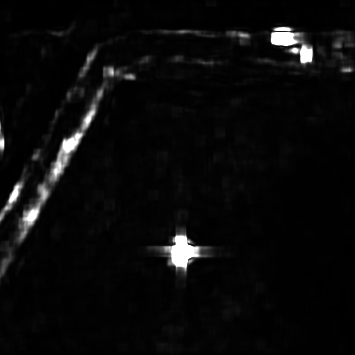}     
  \end{minipage}
  \caption{Comparison of anomaly maps clipped at the highest $2\%$ values: (left) Input image, (middle) $L_1$ loss between $X$ and $\hat{X}$ and (right) $A^{\mathbf{X}}_E$.}
  \label{fig:l1_cov}
\end{figure}

\subsection{Covariance estimation}
Covariance matrices are estimated locally around the pixel under test. This is useful to strongly reduce the noise in the anomaly map compared to a standard $L^1$ loss, as it is shown in Fig.~\ref{fig:l1_cov}. Indeed, to estimate a covariance matrix, we use a sliding window represented by a boxcar $\mathcal{B}_{k,l}$ where $k,l$ is the coordinate of its center. For multivariate Gaussian distribution $\mathcal{N}(\boldsymbol{\mu}, \boldsymbol{\Sigma})$, the Maximum Likelihood Estimators of the mean vector $\boldsymbol{\mu}$ and the covariance matrix $\boldsymbol{\Sigma}$ lead to the well known Sample Mean Vector (SMV) and the Sample Covariance Matrix (SCM) which are defined as:
\begin{equation}
\label{eq:9}
  \hat{\boldsymbol{\mu}}^\mathbf{X}_{k,l}= \displaystyle  \frac{1}{\vert \mathcal{B}_{k,l}\vert } \sum \limits_{i,j \in \mathcal{B}_{k,l}} \mathbf{X}_{i,j}\,,
\end{equation}
\begin{equation}
\label{eq:10}
 \hat{\boldsymbol{\Sigma}}^\mathbf{X}_{k,l} =  \frac{1}{\left|\mathcal{B}_{k,l}\right| } \sum \limits_{i,j \in \mathcal{B}_{k,l}} \left(\mathbf{X}_{i,j} - \hat{\boldsymbol{\mu}}^{\mathbf{X}_{k,l}}\right) \left(\mathbf{X}_{i,j} - \hat{\boldsymbol{\mu}}^{\mathbf{X}_{k,l}}\right)^T \, ,
\end{equation}
where $\hat{\boldsymbol{\mu}}^{\mathbf{X}_{k,l}}$ is the estimate of the mean vector $\boldsymbol{\mu}$, where $\hat{\boldsymbol{\Sigma}}_{k,l}^{\mathbf{X}}$ is the estimate of the covariance matrix $\boldsymbol{\Sigma}$ associated to the boxcar $\mathcal{B}_{k,l}$ in the image $\mathbf{X}$ and where $|.|$ denotes the cardinal operator. 

\begin{figure*}[t]
   \centering
   \includegraphics[width=0.9\linewidth]{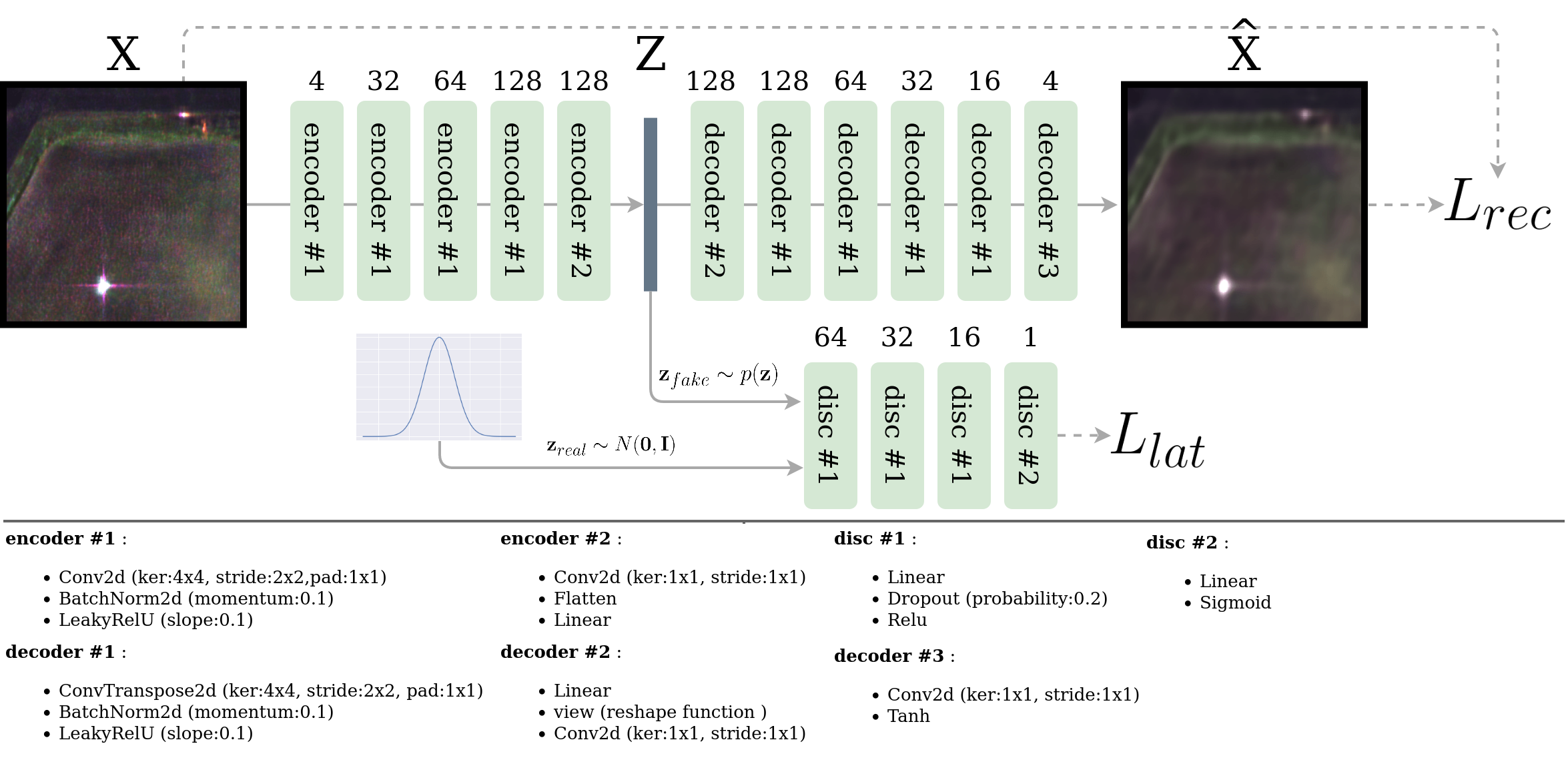}  
  \caption{Architecture of the Adversarial Autoencoder. The numbers above blocks represent the number of input channels for the encoder and the number of output channels for the decoder and the discriminator (disc). The description of each block is done with Pytorch notations.}
  \label{fig:AAE}
\end{figure*}

\subsection{Distance metric}
There are multiple possibilities to compute a distance between two matrices \cite{Forstner03,Dryden09}. A common way to do this consists in computing the square of the Frobenius norm of the difference between the two matrices.
\begin{equation}
\label{eq:11}
A^{\mathbf{X}}_E(k,l) =  \left\|\hat{\boldsymbol{\Sigma}}_{k,l}^{\mathbf{X}} - \hat{\boldsymbol{\Sigma}}_{k,l}^{\hat{\mathbf{X}}}\right\|_F^2 \, .   
\end{equation}

Other methods, based on the Frobenius norm, could also be applied on log-matrices or root matrices.
In our detection case, a lot of importance is given to the intensity difference between $\hat{\boldsymbol{\Sigma}}^{\mathbf{X}}$ and $\hat{\boldsymbol{\Sigma}}^{\hat{\mathbf{X}}}$. The Euclidean metric \eqref{eq:11} highlights the difference in intensity between the two covariance matrices characterizing each pixel and its reconstructed value while preserving a low PFA. The pseudo-code for the proposed anomaly detection method is detailed in Algorithm~\ref{algo:detection}.

\begin{algorithm}[htbp]
\SetAlgoLined
    \PyComment{SAR2SAR: despeckling network} \\
    \PyComment{AAE: reconstruction network} \\
    \PyComment{k: semi kernel size} \\
    \PyComment{scm: compute the sample covariance matrix} \\
    \PyComment{norm: Frobenius norm} \\
    \BlankLine
    \PyCode{for x\_noisy in eval\_data:}\\
    \Indp   
        \PyCode{x = SAR2SAR(x\_noisy)}\\
        \PyCode{x\_rec = AAE(x)}\\
        \PyCode{(\_,m,n) = x.shape} \PyComment{channels,height,width} \\
        \PyCode{ano\_map = zeros(m,n)}\\
        \BlankLine
        \PyComment{Compute anomaly score for each pixel}\\
        \PyCode{for i in range(m):}\\
        \Indp
            \PyCode{for j in range(n):}\\
            \Indp
                \PyComment{Crop to compute scm}\\
                \PyCode{x\_c = x[:,i-k:i+k+1,j-k:j+k+1]}\\
                \PyCode{x\_r\_c = x\_rec[:,i-k:i+k+1,j-k:j+k+1]}\\
                \PyCode{c, c\_r = scm(x\_c),scm(x\_r\_c) }\\
                \PyCode{ano\_map[i,j] = norm(c - c\_r).pow(2)}\\
            \Indm
        \Indm
        \PyCode{m,M = min(ano\_map), max(ano\_map)}\\
        \PyCode{ano\_map = (ano\_map - m) / (M - m)}\\
        \PyCode{save(ano\_map)} \\
    \Indm 
\caption{Anomaly detection pseudocode in python style}
\label{algo:detection}
\end{algorithm}

\section{Experiences and analysis}

In this section, we experiment with the proposed algorithm on the ONERA SETHI SAR dataset. First, the analysis of the despeckling network is presented. We then evaluate the change detection method and compare it with a standard approach.

\subsection{Despeckling quality}

The proposed algorithm is applied to images that are decomposed into patches of size $256\times 256$.

For phase A, we first need to compute a speckle-free image by averaging co-registered images acquired at different dates. Every polarization has been averaged, and we then used the algorithm MuLoG \cite{Deledalle17} to remove the remaining speckle. It gives us training data of 3052 patches grouped in batches of size 32, which leads to a total of 955 batches. The network is then trained for 20 epochs.

Phases B and C use images from the dataset previously described, and the pre-processing step is the one described in Phase B and C, Section \ref{sec:Despeckling}. There are four piles of two images, one for each polarization, and each image is decomposed into 33047 patches with a batch size of 32, which leads to a total of 1032 batches. The network is trained for ten epochs in phases B and C.

\begin{figure}[t]
\centering\begin{minipage}[]{0.42\linewidth}
   \centering
   \includegraphics[width=\linewidth]{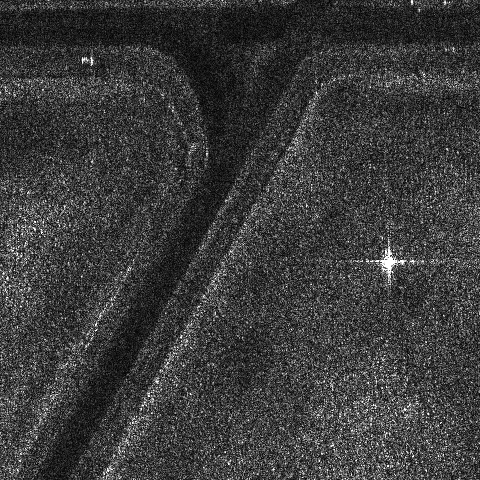}
\end{minipage}
\begin{minipage}[]{0.42\linewidth}
   \centering
   \includegraphics[width=\linewidth]{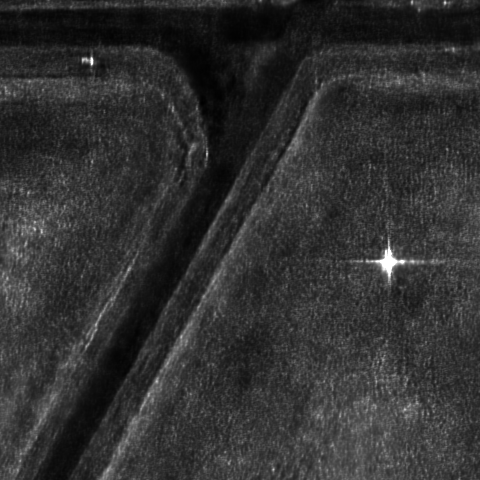} 
  \end{minipage}\vspace{1.5mm} \\
  \begin{minipage}[]{0.42\linewidth}
   \centering
   \includegraphics[width=\linewidth]{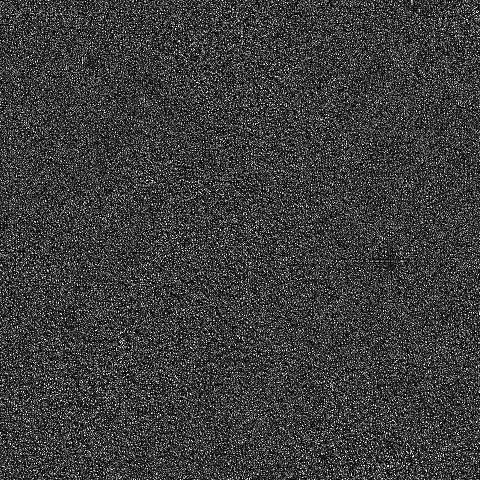} 
  \end{minipage}
  \begin{minipage}[]{0.42\linewidth}
   \includegraphics[width=\linewidth,height=\linewidth]{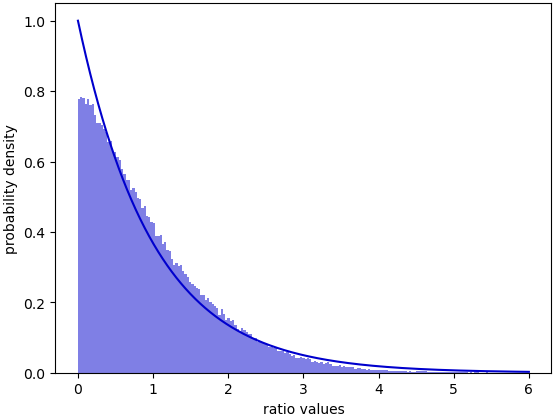}     
  \end{minipage}
  \caption{Comparison between (top left): $HH$ SAR image, (top-right): its speckle-free version. (bottom-left): the intensity ratio of these images. (bottom-right): superposition of the histogram of the ratio image and the theoretical Exponential distribution with parameter $\lambda = 1$.}
  \label{fig:HH_speck}
\end{figure}

Figure~\ref{fig:HH_speck} presents the result of the despeckling process for $HH$ polarization, and the corresponding ratio image is defined as:
\begin{equation}
    ratio = I_{speckle}/I_{despeckling}\, ,
\end{equation}
where $I_{speckle}$ and $I_{despeckling}$ represent respectively the intensity of the original image and the speckle-free image. 

This test image, relatively heterogeneous, comprises vegetation, roads, thin lines, and a strong scatterer annotated as a vehicle. The image dynamic has not been altered, and because there is no structure in the ratio image, we may confidently affirm that our despeckling network has succeeded in only removing the speckle. 

The speckle on its own should follow an exponential distribution with its parameter equal to one. In the histogram in Fig.~\ref{fig:HH_speck}, we found that, as expected, the actual distribution follows an exponential probability density function. 

\subsection{Anomaly detection}

To assess the quality of our results, they are compared with the so-called Reed-Xiaoli detector \cite{Reed90} on complex-valued SAR images. The algorithm consists in estimating locally, around a pixel characterized by its polarimetric response $\bf{x}$, the associated mean vector $\hat{\boldsymbol{\mu}}$ and the covariance matrix $\hat{\boldsymbol{\Sigma}}$ of its surrounding background and then to test if this pixel under test is belonging, or not, to this background. The anomaly score is computed through the well-known Mahalanobis distance:
\begin{equation}
\label{eq:rx}
RX(\bf{x}) = \left(\bf{x}-\hat{\boldsymbol{\mu}}\right)^H\,\hat{\boldsymbol{\Sigma}}^{-1}\,\left(\bf{x}-\hat{\boldsymbol{\mu}}\right)\, ,  
\end{equation}
where $H$ denotes the transpose conjugate operator. The parameters are locally estimated through the Gaussian Sample Mean Vector and Sample Covariance Matrix estimates. An exclusion window prevents the use of anomalous data in parameter estimation (guard cells). 

Additional comparisons will be made using our proposed AAE algorithm without the despeckling pre-processing and the $L^1$ norm between $\mathbf{X}$ and $\hat{\mathbf{X}}$.

To run the proposed algorithm, we first need to train the AAE on the dataset described in \ref{section:SAR}.\ref{subsection:data}

\subsubsection{AAE training and architecture}

The training process is self-supervised. The training dataset has no labels and potentially can contain anomalies. In this case, the evaluation and training datasets can be mixed because the training process requires only unlabeled data. In the same way as for the despeckling algorithm, a log transformation is used, and the input data are normalized between 0 and 1. These images are decomposed into patches of size $64\times 64$. The sliding window has a stride of 16. This leads to a total of 163785 patches grouped in batches of size 128, so we get 1279 batches for one training epoch with a total of 20 epochs.

The architecture of the network is illustrated in Fig.~\ref{fig:AAE}. It has been designed according to \cite{Radford16}. To update the weights, we use the optimizer Adam \cite{Kingma17} with a cyclical learning rate \cite{Smith15} that goes linearly from $10^{-3}$ to $10^{-2}$. It takes 2558 batches to go from one value to another. A complete cycle will take four epochs. This method helps to have a robust training phase that will converge even if we do not know the perfect learning rate for the network.

\subsubsection{Evaluation dataset}

The proposed method has been qualitatively evaluated on a known area of the dataset described in Section \ref{section:SAR}.\ref{subsection:data}. Since there is no ground truth available, the boundary between the anomaly and the background is not apparent, which makes annotation almost impossible in many cases. The area is composed of known anomalies (vehicles) intentionally placed by the ONERA team during a measurement campaign.

\begin{figure}[t]
\begin{minipage}[]{0.325\linewidth}
   \centering
        \small{(a)}   
  \end{minipage}
  \begin{minipage}[]{0.325\linewidth}
   \centering
        \small{(b)}   
  \end{minipage}
  \begin{minipage}[]{0.325\linewidth}
   \centering
        \small{(c)}   
  \end{minipage}\\
\begin{minipage}[]{0.02\linewidth}
   \centering
   \rotatebox{90}{\small{Noisy}}    
  \end{minipage}
\begin{minipage}[]{0.31\linewidth}
   \centering
   \includegraphics[width=\linewidth]{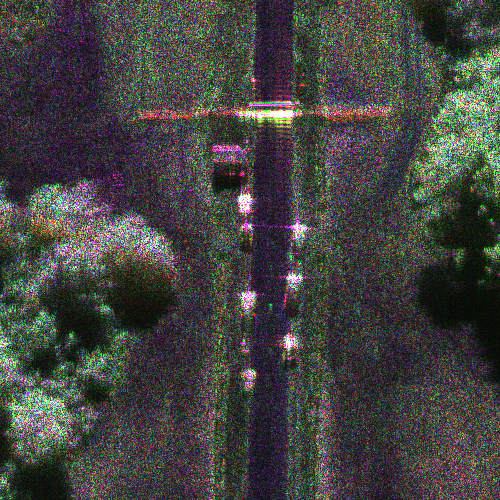}     
  \end{minipage}
    \hfill
  \begin{minipage}[]{0.31\linewidth}
   \centering
   \includegraphics[width=\linewidth]{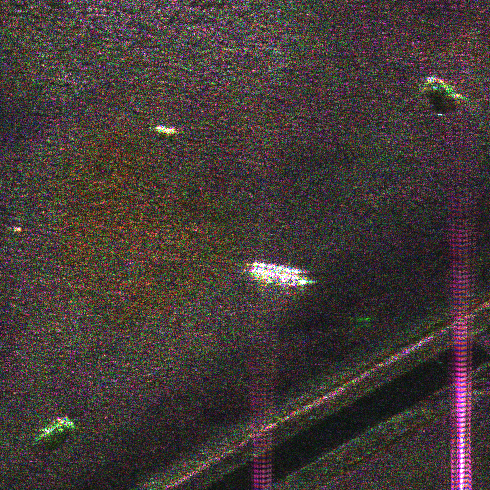}     
  \end{minipage}
  \hfill
  \begin{minipage}[]{0.31\linewidth}
   \centering
   \includegraphics[width=\linewidth]{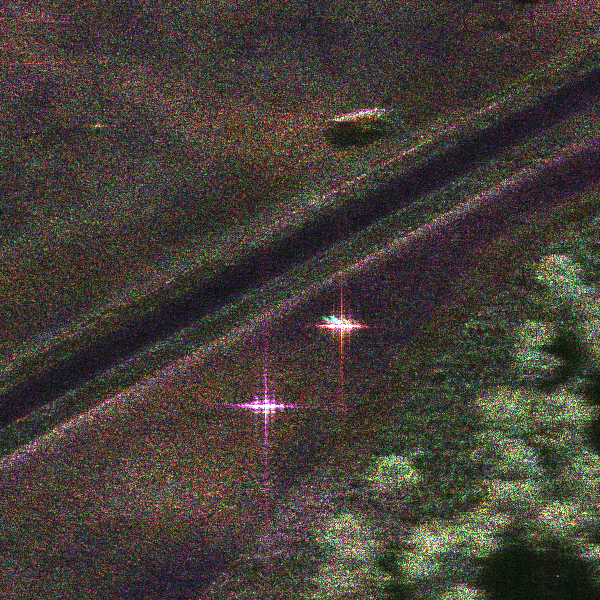}     
  \end{minipage}\vspace{1.5mm}\\
  
  \begin{minipage}[]{0.02\linewidth}
   \centering
   \rotatebox{90}{\small{Denoised}}    
  \end{minipage}
\begin{minipage}[]{0.31\linewidth}
   \centering
   \includegraphics[width=\linewidth]{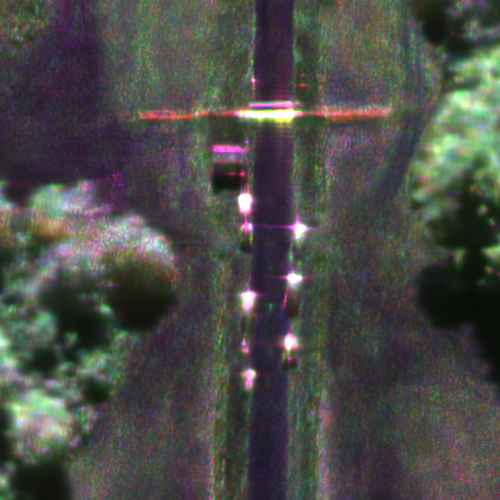}     
  \end{minipage}
    \hfill
  \begin{minipage}[]{0.31\linewidth}
   \centering
   \includegraphics[width=\linewidth]{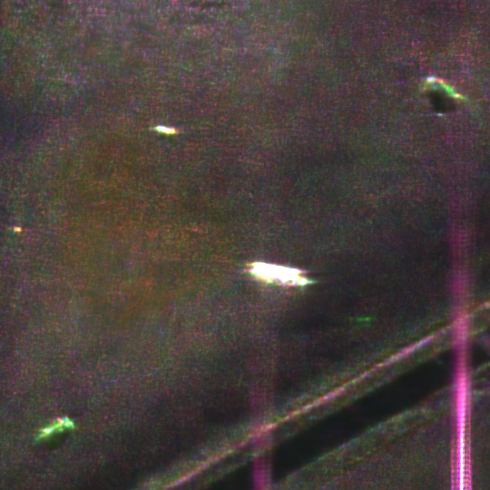}     
  \end{minipage}
  \hfill
  \begin{minipage}[]{0.31\linewidth}
   \centering
   \includegraphics[width=\linewidth]{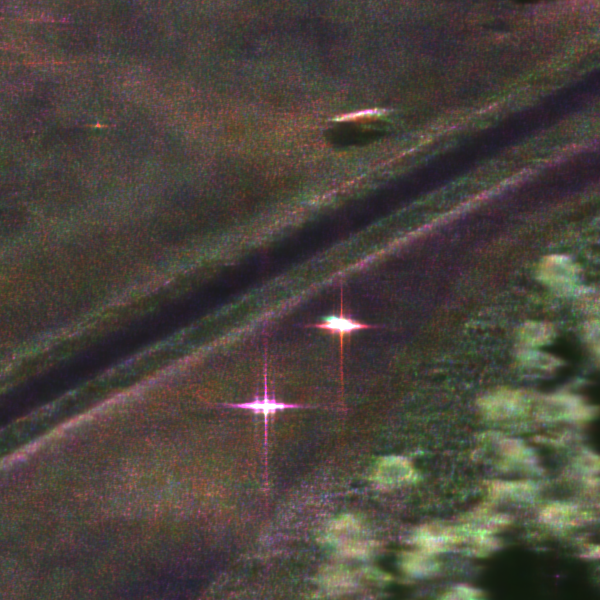}     
  \end{minipage}\vspace{1.5mm}\\
  
  \begin{minipage}[]{0.02\linewidth}
   \centering
   \rotatebox{90}{\small{Reconstruction}}    
  \end{minipage}
\begin{minipage}[]{0.31\linewidth}
   \centering
   \includegraphics[width=\linewidth]{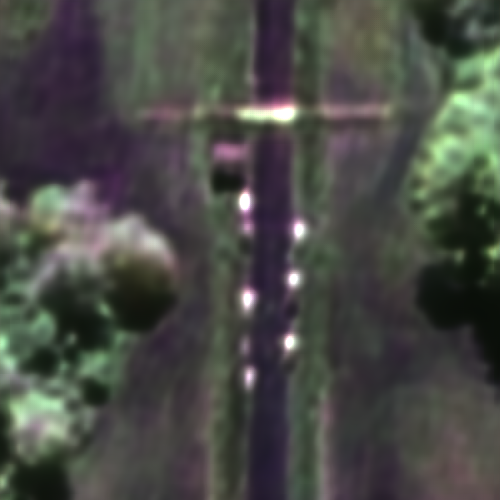}     
  \end{minipage}
    \hfill
  \begin{minipage}[]{0.31\linewidth}
   \centering
   \includegraphics[width=\linewidth]{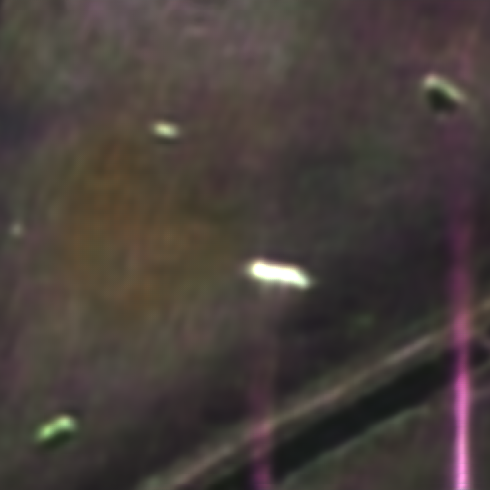}     
  \end{minipage}
  \hfill
  \begin{minipage}[]{0.31\linewidth}
   \centering
   \includegraphics[width=\linewidth]{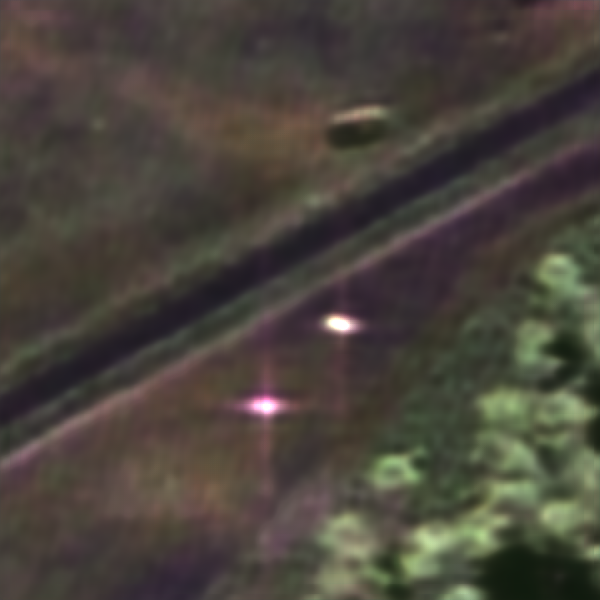}    
   \end{minipage}\vspace{1.5mm}\\
   
   \begin{minipage}[]{0.02\linewidth}
   \centering
   \rotatebox{90}{\small{Difference}}    
  \end{minipage}
\begin{minipage}[]{0.31\linewidth}
   \centering
   \includegraphics[width=\linewidth]{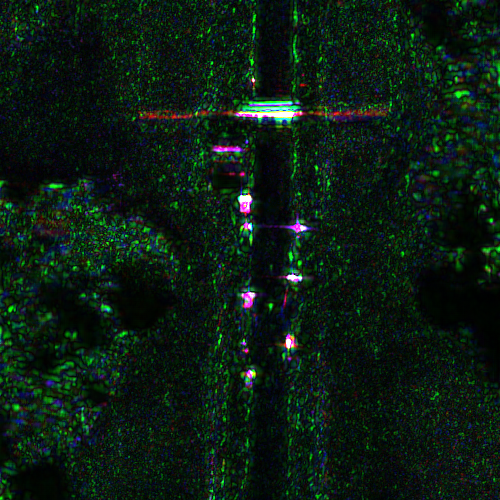}     
  \end{minipage}
    \hfill
  \begin{minipage}[]{0.31\linewidth}
   \centering
   \includegraphics[width=\linewidth]{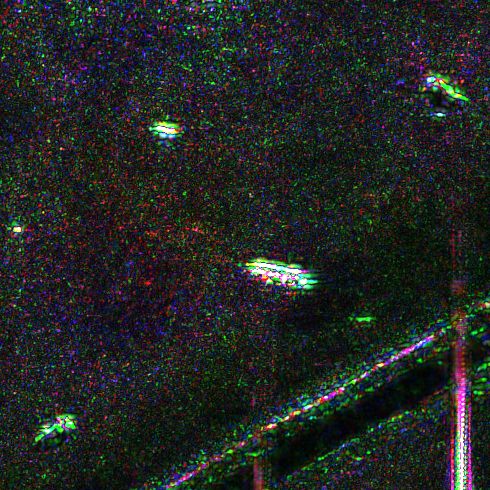}     
  \end{minipage}
  \hfill
  \begin{minipage}[]{0.31\linewidth}
   \centering
   \includegraphics[width=\linewidth]{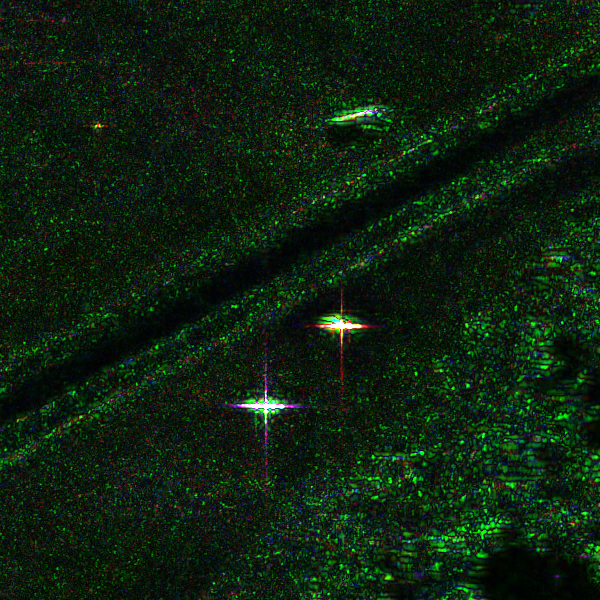}    
   \end{minipage}
   
  \caption{Evolution of the different steps of the process for three different areas of the SAR image. (left): (a) SAR image containing a fence and vehicles. (b): SAR image containing vehicles and trihedral sidelobes corresponding to purple lines. (c): SAR image containing three vehicles.}
  \label{fig:input}
\end{figure}

\begin{figure}[]
\centering
    \begin{minipage}[]{0.67\linewidth}
    \centering
    \hspace{0.1\linewidth}
    \small{HH}   
  \end{minipage}\vspace{0.5mm}\\
  \begin{minipage}[]{0.02\linewidth}
   \centering
   \rotatebox{90}{\small{Intensity}}    
  \end{minipage}
  \begin{minipage}[]{0.67\linewidth}
   \centering
   \includegraphics[width=\linewidth]{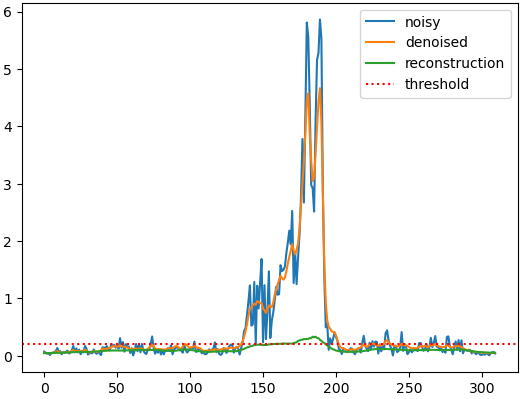}    
  \end{minipage}\vspace{0.5mm}\\
  \begin{minipage}[]{0.67\linewidth}
   \centering
   \hspace{3mm}
   \small{Position}    
  \end{minipage}\vspace{1mm}\\
  
\begin{minipage}[]{0.67\linewidth}
   \centering
    \hspace{0.1\linewidth}
    \small{HV}   
  \end{minipage}\vspace{0.5mm}\\
  \begin{minipage}[]{0.02\linewidth}
   \centering
   \rotatebox{90}{\small{Intensity}}    
  \end{minipage}
  \begin{minipage}[]{0.67\linewidth}
   \centering
   \includegraphics[width=\linewidth]{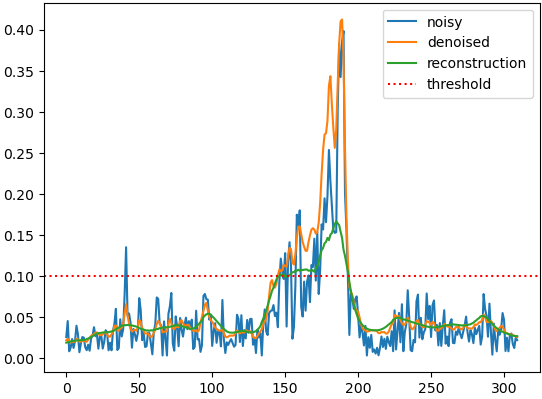}     
  \end{minipage}\vspace{0.5mm}\\
  \begin{minipage}[]{0.67\linewidth}
   \centering
   \hspace{3mm}
   \small{Position}    
  \end{minipage}
  
    \caption{Comparisons of horizontal profiles on a strong scatterer of the image (a) in Figure \ref{fig:input} in $HH$ and $HV$ polarization. }
    \label{fig:coupe}
\end{figure}

\subsubsection{Qualitative results}

The evaluation of noisy images is displayed in Fig.~\ref{fig:input} with their denoised and reconstructed versions. The map named {\it Difference} represents the absolute value between
Denoised and Reconstruction maps. We can remark that the reconstructed images are blurry, but all the essential structures are kept. 

The despeckling pre-processing is not removing any structure, even the smallest one, like the green line in the bottom right of the image (b). In all the reconstructed images, the intensity of each point-like anomaly is greatly reduced. Figure~\ref{fig:coupe} illustrates this fact and shows a comparison of horizontal profiles at the location characterizing a strong scatterer of image (a). Red dotted lines correspond to the visualization threshold defined in Eq.~\eqref{eq:4}. This intensity attenuation is fundamental to the algorithm because the change detection is only applied between the denoised and the reconstructed image in order to highlight potential anomalies.

\begin{figure}[]
\begin{minipage}[]{0.04\linewidth}
   \centering
   \rotatebox{90}{\small{$A^{\mathbf{X}}_E$}}    
  \end{minipage}
\begin{minipage}[]{0.31\linewidth}
   \centering
   \includegraphics[width=\linewidth]{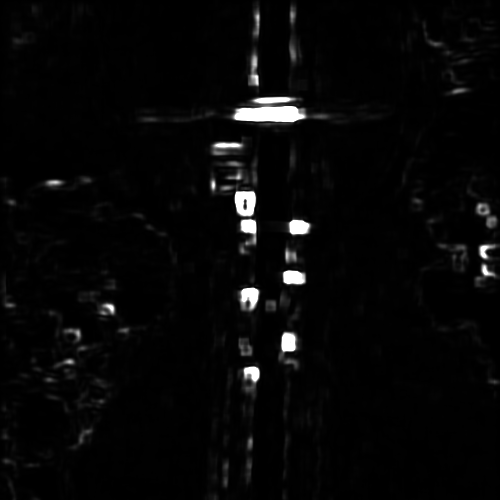}     
  \end{minipage}
    \hfill
  \begin{minipage}[]{0.31\linewidth}
   \centering
   \includegraphics[width=\linewidth]{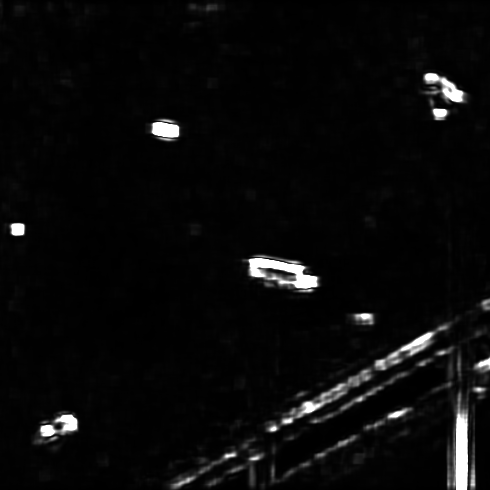}     
  \end{minipage}
  \hfill
  \begin{minipage}[]{0.31\linewidth}
   \centering
   \includegraphics[width=\linewidth]{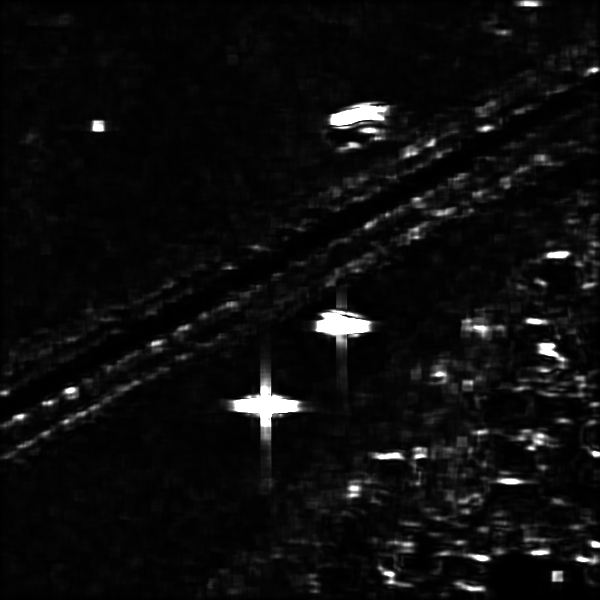}     
  \end{minipage}\vspace{0.8mm}\\

\begin{minipage}[]{0.04\linewidth}
   \centering
   \rotatebox{90}{\small{RX}}    
  \end{minipage}
\begin{minipage}[]{0.3073\linewidth}
   \centering
   \includegraphics[width=\linewidth]{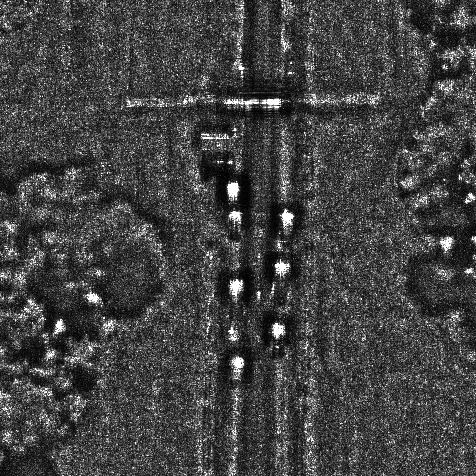}     
  \end{minipage}
    \hfill
  \begin{minipage}[]{0.3073\linewidth}
   \centering
   \includegraphics[width=\linewidth]{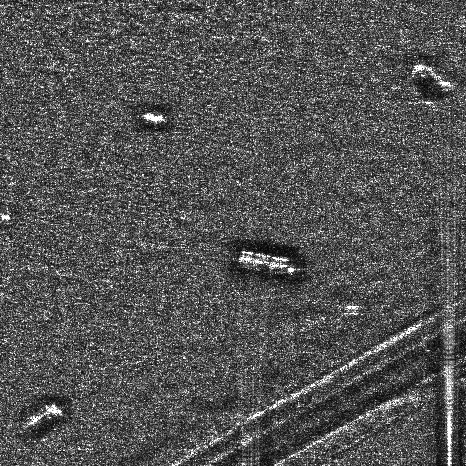}     
  \end{minipage}
  \hfill
  \begin{minipage}[]{0.3073\linewidth}
   \centering
   \includegraphics[width=\linewidth]{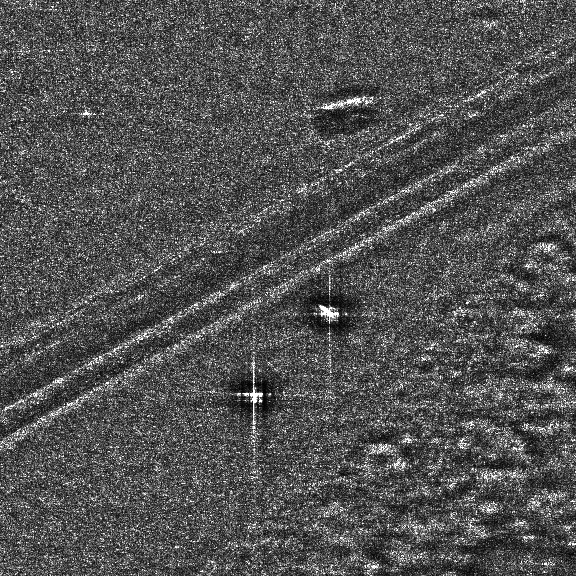} 
  \end{minipage}\vspace{0.8mm}\\
  
  \begin{minipage}[]{0.04\linewidth}
   \centering
   \rotatebox{90}{\small{$L^1(\mathbf{X},\hat{\mathbf{X}})$}}    
  \end{minipage}
\begin{minipage}[]{0.31\linewidth}
   \centering
   \includegraphics[width=\linewidth]{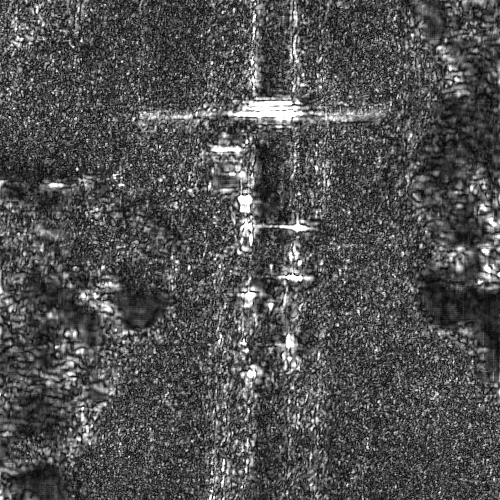}     
  \end{minipage}
    \hfill
  \begin{minipage}[]{0.31\linewidth}
   \centering
   \includegraphics[width=\linewidth]{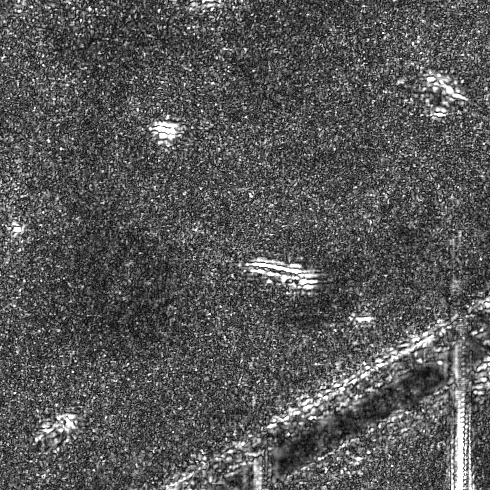}     
  \end{minipage}
  \hfill
  \begin{minipage}[]{0.31\linewidth}
   \centering
   \includegraphics[width=\linewidth]{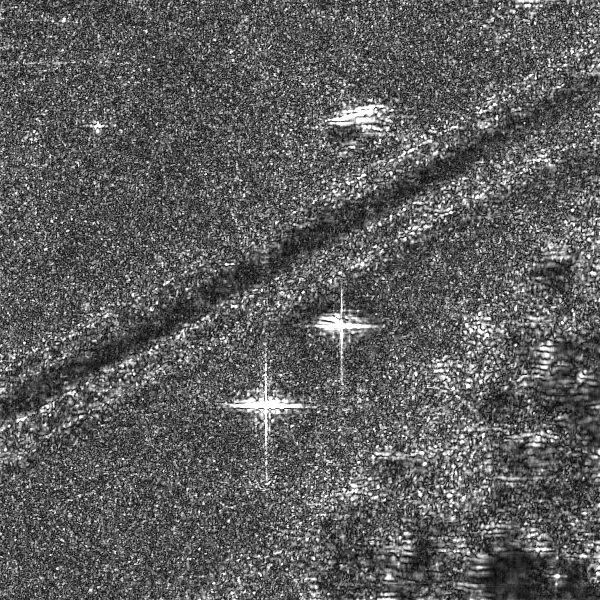}     
  \end{minipage}\vspace{0.8mm}\\

    \begin{minipage}[]{0.04\linewidth}
   \centering
   \rotatebox{90}{\small{$A^{\mathbf{X_{noisy}}}_E$}}    
  \end{minipage}
\begin{minipage}[]{0.31\linewidth}
   \centering
   \includegraphics[width=\linewidth]{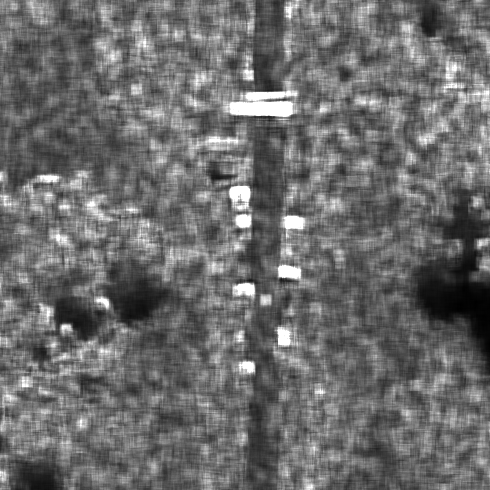}     
  \end{minipage}
    \hfill
  \begin{minipage}[]{0.31\linewidth}
   \centering
   \includegraphics[width=\linewidth]{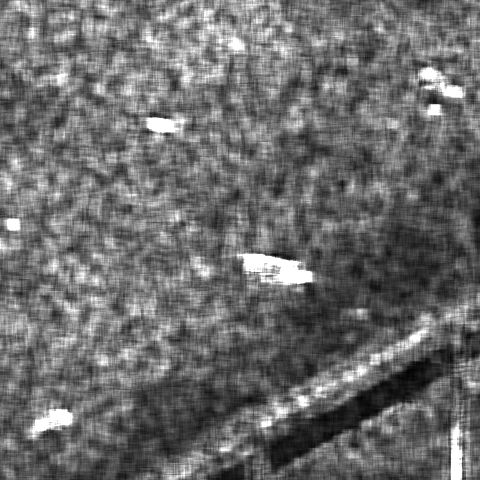}     
  \end{minipage}
  \hfill
  \begin{minipage}[]{0.31\linewidth}
   \centering
   \includegraphics[width=\linewidth]{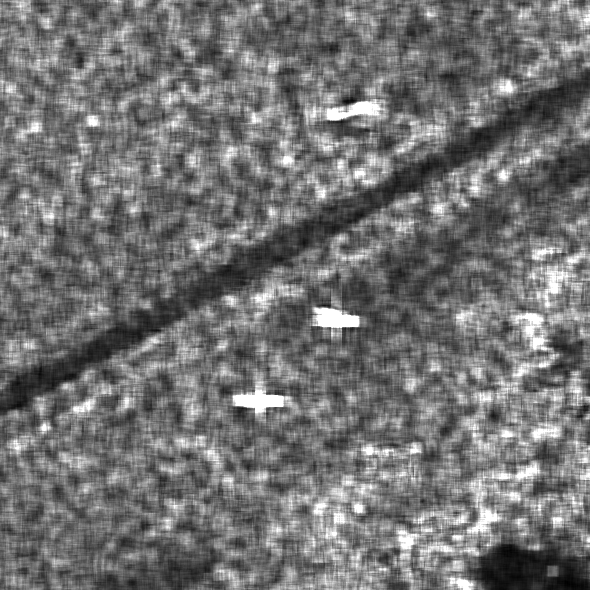}     
  \end{minipage}\vspace{0.8mm}\\
  
  \caption{Comparison between different anomaly maps obtained from the images described in Fig.~\ref{fig:input}.}
  \label{fig:res}
\end{figure}

 Figure~\ref{fig:res} presents a comparison of anomaly detection results obtained with different methods. All the results are here clipped at the highest $1\%$ values, see Eq~\ref{eq:PFA}.
 
 The anomaly maps obtained with the methods $A^{\mathbf{X_{noisy}}}_E$, the Reed-Xiaoli RX detector and  $L^1(\mathbf{X},\hat{\mathbf{X}})$ have a similar drawback characterizing the presence of a large number of false detections. The proposed method $A^{\mathbf{X}}_E$ is shown to have better performance for the same PFA. 

\begin{figure*}[t]
\begin{minipage}[]{0.16\linewidth}
   \centering
        \small{input}   
  \end{minipage}
  \begin{minipage}[]{0.16\linewidth}
   \centering
        \small{label}   
  \end{minipage}
  \begin{minipage}[]{0.16\linewidth}
   \centering
   \small{RX} 
  \end{minipage}
  \begin{minipage}[]{0.16\linewidth}
   \centering
        \small{$A^{\mathbf{X_{noisy}}}_E$}  
  \end{minipage} 
  \begin{minipage}[]{0.16\linewidth}
   \centering
   \small{$L^1(\mathbf{X},\hat{\mathbf{X}})$}
  \end{minipage}
  \begin{minipage}[]{0.16\linewidth}
   \centering
        \small{$A^{\mathbf{X}}_E$}
  \end{minipage}\vspace{1mm}\\
\begin{minipage}[]{0.16\linewidth}
   \centering
   \includegraphics[width=\linewidth]{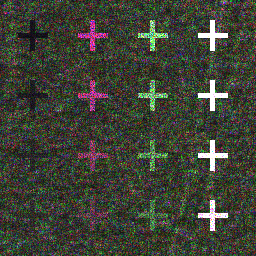}     
  \end{minipage}
  \begin{minipage}[]{0.16\linewidth}
   \centering
   \includegraphics[width=\linewidth]{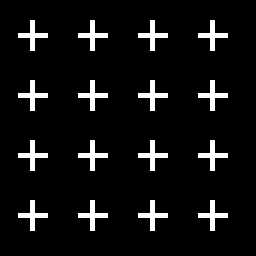}     
  \end{minipage}
  \begin{minipage}[]{0.16\linewidth}
   \centering
   \includegraphics[width=\linewidth]{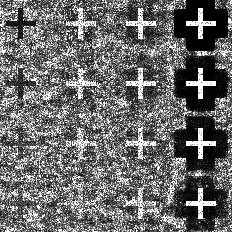}     
  \end{minipage}
 \begin{minipage}[]{0.16\linewidth}
   \centering
   \includegraphics[width=\linewidth]{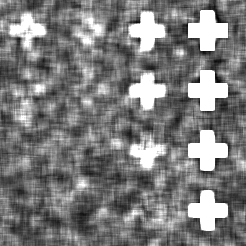}     
  \end{minipage}
  \begin{minipage}[]{0.16\linewidth}
   \centering
   \includegraphics[width=\linewidth]{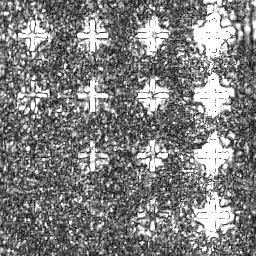}     
  \end{minipage} 
  \begin{minipage}[]{0.16\linewidth}
   \centering
   \includegraphics[width=\linewidth]{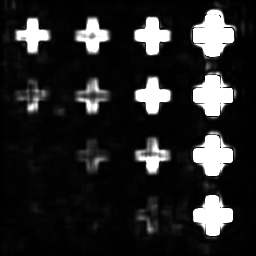}     
  \end{minipage} 
  \caption{Comparison between different anomaly maps on a real SAR image  with fake anomalies and its corresponding label.}
  \label{fig:res_anomaly}
\end{figure*}

\subsubsection{Quantitative results}
For quantitative evaluation, we have embedded synthetic test patterns with different intensity levels in a true anomaly-free crop SAR image. 

\begin{figure}
   \centering
   \input{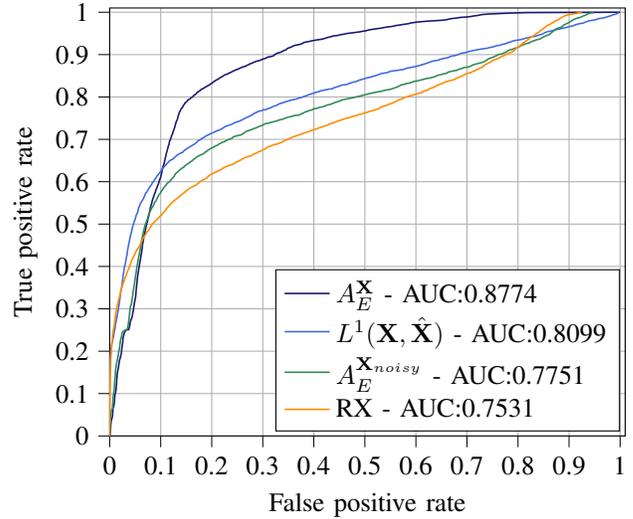}
  \caption{ROC curve and value of the AUC for the anomaly detection algorithms displayed Fig.~\ref{fig:res_anomaly}.}
  \label{fig:res_fake}
\end{figure}

The fake anomaly map results are represented in Fig.~\ref{fig:res_anomaly}. The values are based on true intensities characterizing the dataset with a decreasing significance from top to bottom. In this figure, anomaly detection maps are clipped at $10\%$ of the higher value because some anomalies have a value close to the background. This makes the detection harder; thus, we reduce the dynamic range to have a better representation. One advantage of autoencoder-based detection is detecting abnormal areas of low intensities (left cross), contrary to the RX detector. The proposed $A^{\mathbf{X}}_E$ method outperforms all the other ones and even for the detection of low-intensity patterns. This is illustrated in Fig.~\ref{fig:res_fake}, which shows the overall performance obtained in terms of Area Under the Curve (AUC). The proposed $A^{\mathbf{X}}_E$ still has the better performance. 

\section{Conclusion}
In this article, we propose a novel anomaly detection algorithm. It is mainly based on the use of adversarial autoencoders. They have not been widely used in the SAR community because of the speckle noise, which dramatically degrades algorithm performance. Thanks to the recent advances in deep learning despeckling algorithms specifically developed for SAR images, we can now efficiently develop new algorithms based on these methods to enhance anomaly detection performance. 

Our proposed algorithm outperforms the conventional Reed-Xiaoli method since it can detect abnormal areas with low-intensity values. Because this self-supervised strategy does not require labeled data, it can easily be extended to another type of data as long as the anomaly quantity remains negligible. 

The perspectives envisaged concern the improvement of change detection techniques that can help in giving better detection performance as well as in regulating the False Alarm Rate.

\end{document}